\documentclass[letterpaper, 10 pt, conference]{ieeeconf}  

\IEEEoverridecommandlockouts                              

\usepackage{graphics} 
\usepackage{epsfig} 
\usepackage{hyperref}
\usepackage{mathptmx} 
\usepackage{times} 
\usepackage{amsmath} 
\usepackage{amssymb}  
\usepackage{todonotes}
\usepackage{booktabs}
\usepackage{multicol}
\usepackage{multirow}
\graphicspath{{image/}}
\usepackage{todonotes}
\usepackage{caption}
\usepackage{soul}
\usepackage{svg}
\usepackage{float}
\usepackage{afterpage} 
\definecolor{mycolor}{HTML}{BBCCBA}
\usepackage{svg}
\let\labelindent\relax
\usepackage{enumitem}

\definecolor{OrangeRed}{HTML}{FF4500}
\definecolor{DodgerBlue}{HTML}{1E90FF}
\definecolor{MediumSeaGreen}{HTML}{3CB371}
\definecolor{Gold}{HTML}{B8860B}

\title{\LARGE \bf
PANOS: Payload-Aware Navigation in Offroad Scenarios
}
\newcommand{\insertfig}{
\includegraphics[trim=1.25cm 1.25cm 1.25cm 1.25cm, clip=true,width=\textwidth]{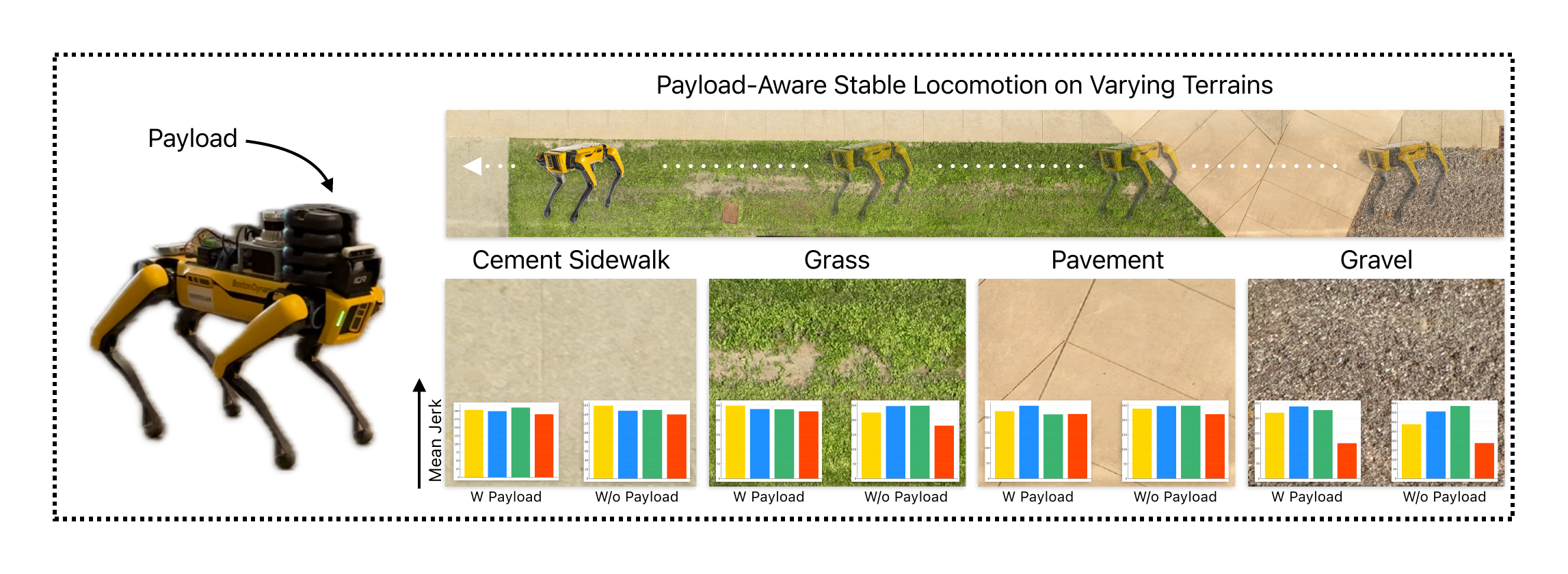}
\captionof{figure}{Stable navigation using PANOS velocity estimations under varying payloads on diverse terrains. In the sub-graphs, we compare the stabilities (Mean Jerk $\bar{J}$ acting on the robot's body) estimated from \textcolor{Gold}{Spot's inbuilt}~\cite{bostondynamics2023} sensing system, \textcolor{DodgerBlue}{VAPOR}~\cite{weerakoon2024vapor}, \textcolor{MediumSeaGreen}{VERN}~\cite{sathyamoorthy2023vern}, and \textcolor{OrangeRed}{PANOS} (From Left to Right respectively). We determine the stabilities W and W/o Payloads by measuring the average jerk due to the texture-rich terrains. 
}\vspace{-0.5cm}
\label{fig:firstpage_intro}
}
\makeatletter
\apptocmd{\@maketitle}{\centering\insertfig}{}{}
\makeatother

\author{Kartikeya Singh, Yash Turkar, Christo Aluckal, Charuvarahan Adhivarahan and Karthik Dantu
\thanks{All authors are from the Department of Computer Science and Engineering.
        University at Buffalo, Buffalo, NY (emails:{ksingh35,yashturk,christoa,charuvah, kdantu} @buffalo.edu)
        }%
\thanks{Corresponding Author: ksingh35@buffalo.edu}
\thanks{\textcolor{blue}{\href{https://github.com/kartikeya13/IMPACT}{Project Page}}}
\thanks{The authors wish to thank and acknowledge partial support
from grants NSF-1846320, ONR N00014-23-S-C001 as well as a gift from Moog Inc.}
}
\begin{document}
\maketitle
\thispagestyle{empty}
\pagestyle{empty}


\begin{abstract} 
Nature has evolved humans to walk on different terrains by developing a detailed understanding of their physical characteristics. Similarly, legged robots need to develop their capability to walk on complex terrains with a variety of task-dependent payloads to achieve their goals. However, conventional terrain adaptation methods are susceptible to failure with varying payloads. In this work, we introduce PANOS, a weakly supervised approach that integrates proprioception and exteroception from onboard sensing to achieve a stable gait while walking by a legged robot over various terrains. Our work also provides evidence of its adaptability over varying payloads. We evaluate our method on multiple terrains and payloads using a legged robot. PANOS improves the stability up to \textbf{44\%} without any payload and \textbf{53\%} with 15 lbs payload. We also notice a reduction in the vibration cost of \textbf{20\%} with payload for various terrain types when compared to state-of-the-art methods.

\end{abstract}


\section{INTRODUCTION}\label{intro}

Autonomous navigation of legged robots through diverse and complicated environments has gained attention over the past few years to accomplish critical tasks like payload delivery~\cite{figliozzi2020autonomous}, search and rescue~\cite{bellicoso2018advances}, \cite{eich2008versatile}, environmental inspection~\cite{kolvenbach2019haptic}, agricultural tasks~\cite{naik2016precision} and others. Legged robots like Boston Dynamic's SPOT, Unitree Go1/Go2, and Anybotics ANYmal provide basic legged locomotion, enabling them to walk on different terrains. The primary challenge in using them for offroad navigation is to incorporate a detailed understanding of the terrain and use this for high-level planning and control for stable locomotion.


Several approaches in the past have proposed methods to identify rich semantic information of an off-road terrain by using images~\cite{viswanath2021offseg}, \cite{guanganav}, \cite{zhong2022off}, \cite{yang2023wait}. These methods are capable of identifying semantic class distributions across distinct offroad environments but are not robust to illumination and viewpoint changes. To overcome these issues, prior works make use of LiDAR point clouds~\cite{viswanath2023off}, \cite{dabbiru2020lidar} that remain unperturbed in varying lighting conditions. Additionally, diffusion using various high-level sensors~\cite{kim2024ufo}, \cite{gao2021fine} results in a more detailed representation of the surfaces. These exteroception-based approaches are robust to the segmentations of fine-grained terrains but do not account for the control/planning and stable locomotion of a robot (specifically, legged robots) on these segmented classes. 

Sim-to-real approaches~\cite{agarwal2023legged}, \cite{fu2022coupling} have been explored in establishing twin policies to make the robot learn \textit{Appearance} and \textit{Tactility} about the terrain type explicitly. But these approaches are tightly coupled with the simulated representations they acquire for training (like scan-dots in~\cite{agarwal2023legged}). These simulated-experience learning approaches are well crafted to adapt the physical attributes like agility, traction control, or dexterity to perform control and coordination tasks but don't adapt to real-world nuances such as variable payloads and stability. In this work, we present an approach that learns from real-world experiences of traversal of various terrains instead of adapting a simulated representation. 

\begin{figure*}
    \centering
    \includegraphics[width=\textwidth]{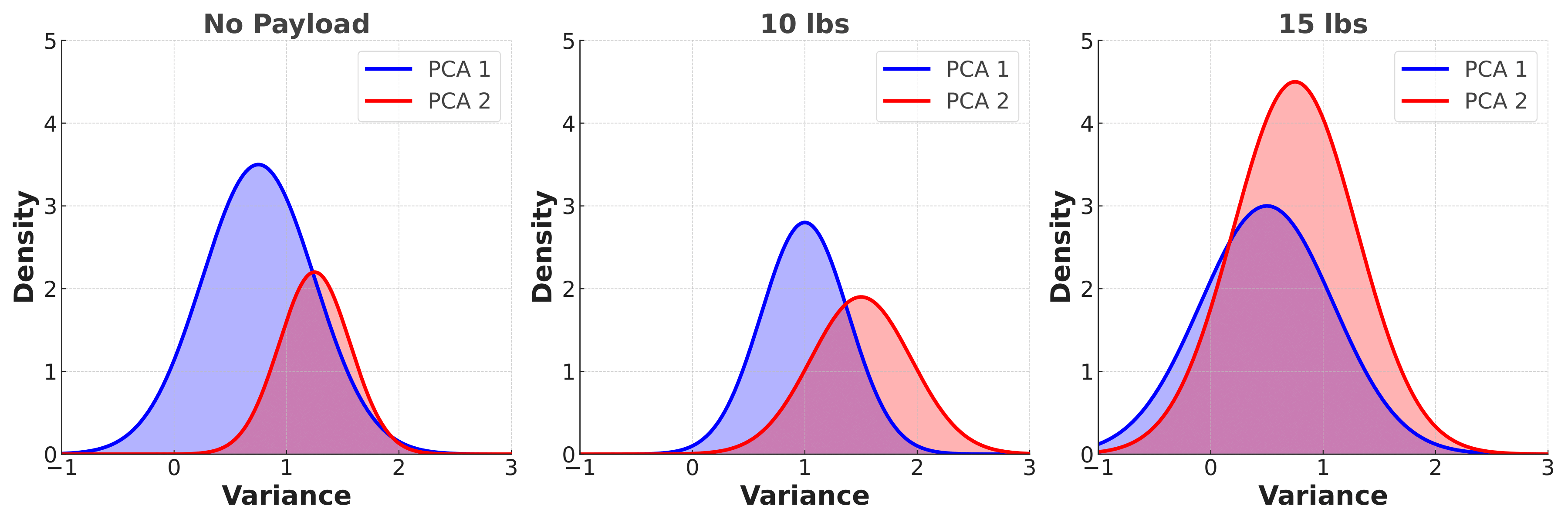}
    \caption {Variance plots from PCA-based measure of proprioception on the same terrain with varying payloads. We can observe the uncertainty from these plots when characterizing the terrain. Therefore, PANOS unwraps the proprioceptive measures and establishes a one-to-one contextual relationship between the proprioception and terrain type (using exteroception).
    }
    \label{fig:pca}
    \vspace{-10pt}
\end{figure*}

Methods using proprioception (velocity/hip positions) from the robot's locomotion along with exteroception from distinct high-level sensors(camera/lidar) have also been used for terrain adaptation. Prior works~\cite {sathyamoorthy2023vern}, \cite{sathyamoorthy2023using}, \cite{weerakoon2023adventr}, \cite{weerakoon2022graspe} have relied solely on exteroception to do planning and navigation of legged robots over complex environments. We conjecture that stable locomotion could be more robustly determined by coupling information from exteroception and proprioception~\cite{elnoor2024pronav}, \cite{weerakoon2024vapor}, \cite{agarwal2023legged}. These approaches not only estimate the terrain representation but also guide a legged robot in terms of regulated velocities and adapt to a suitable state (i.e. Gait switching between crawl/walk/trot or holonomic/non-holonomic motion). However, none of these approaches are robust to changes in the payload mounted on the robot. Any such change will likely need retraining of their models and will work only for the trained payload. In this work, we acquire the robot's state with varying payloads to estimate the desired velocities in order to provide adequate stability to the robot while traversing through offroad terrains. Prior work~\cite{weerakoon2024vapor}, \cite{elnoor2024pronav} used a reduced representation of proprioception (PCA at various joints) as a measure of stability. This measure works well to achieve safe locomotion. However, changes in payload lead to varying values (See~\autoref{fig:pca}). Such variation demonstrates the possibility of reduced stability if the robot is carrying a different payload than what the model was trained on. Our primary observation of stability is related not just to the nature of the terrain but also to changes in payload on the robot. This is a common occurrence with changes in sensors, computing, or other things the robots carry based on the task at hand. 

In this work, we develop PANOS (\autoref{fig:firstpage_intro}) - a payload-influenced velocity estimation method through a weakly supervised model that is capable of adapting navigation velocity with changes in the payload of the robot. The architecture introduced in this paper selects the most context-aware sequences (inspired from language-based sequential modeling~\cite{vinyals2015pointer}, \cite{min2023recent}) that represent terrain type (using exteroception) with its corresponding proprioception (joints, hips, feet) for this purpose. 

The main contributions of this work are as follows:
\begin{itemize}[leftmargin=1em,topsep=0.1em,itemsep=0.05em]
  \item  We introduce PANOS, a learning-based velocity estimation method that explicitly incorporates terrain type (using camera-based exteroception) and the robot's payload (through proprioception). 
  \item Our framework provides robustness to the velocity estimation over different terrains. It also adapts the learning parameters accounting to the robot's stability even with different payloads.
  \item From our evaluations, PANOS improves the stability of a legged robot by \textbf{44\%} without any payload and \textbf{53\%} with a 15 lbs payload when traversed through various offroad terrains. Our method also reduces the joint vibrations by \textbf{20\%} when compared with two state-of-the-art approaches and the robot's inbuilt locomotion system.
\end{itemize}

\begin{figure*}
    \centering
    \includegraphics[width=\textwidth]{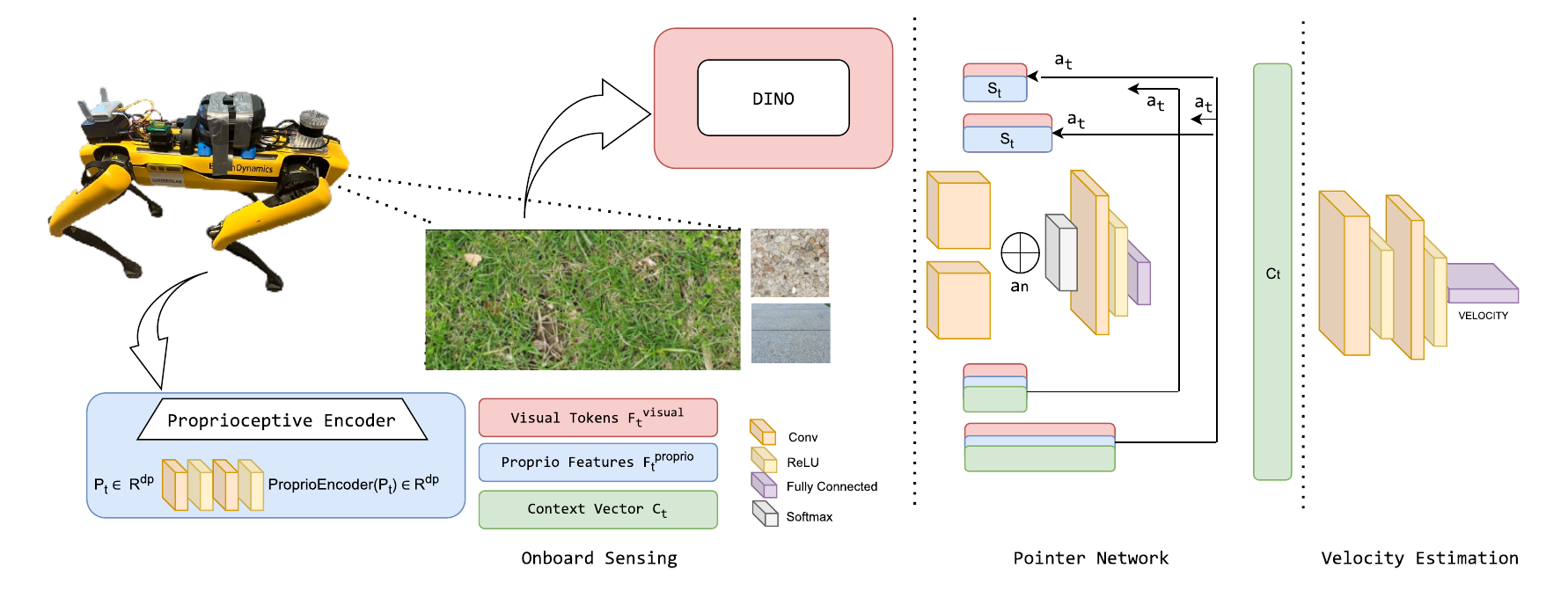}
    \caption{\textbf{Overview of the Pipeline:} PANOS inputs a stream of images and proprioception data $\mathbf{P}_t$ (joints, hips, and, feet slips) recorded in an unsupervised fashion. The framework encodes these readings into two backbones DINOv2~\cite{oquab2023dinov2} and a vanilla encoder for proprioception resulting in sets of random sequences $S_t$ with visual tokens $\mathbf{F}_t^{\text{visual}}$ and proprioceptive features $\mathbf{F}_t^{\text{proprio}}$ stacked together. As an intermediate step, a pointer network is defined to assign the weighted confidence $\text{Confidence}_t$ between these sets and select the dominating ones to train. Finally, we use the trained contextual relationship $\mathbf{c}_t$ as the input to a neural network that predicts the optimal velocity.} \vspace{-0.5cm}
    \label{overview}
\end{figure*}

\section{Related Work}\label{sec:related_work}

\subsection{Terrain Estimation Using Sensing}
Legged robot's navigation capability has been validated~\cite{wermelinger2016navigation}, \cite{agarwal2023legged} across complex, uneven, and varying terrain while ensuring safety and efficiency. With the hardware advancements in the industry, these robots have not been left untouched from performing some of the extreme navigation tasks like ~\cite{wellhausen2023artplanner}, \cite{fu2022coupling}. Perpetual observation of terrain is a vital step to make a robust and collision-free navigation of a robot successful. Sensor-based perception using exteroception has been exploited over the years that could result in a high-level but coursed characterization of a specific terrain. 

Leveraging sensor-based perception works like~\cite{sathyamoorthy2023vern}, \cite{sathyamoorthy2024convoi} utilizes images to determine a risk-aware path planning for a legged robot. However, these methods without any additional signal result in a lack of internal feedback and could make the robot's body unaware of its current state. To overcome these issues,~\cite{elnoor2024pronav} couple both exteroceptive and proprioceptive parameters of a robot to adapt between different gaits. They utilize PCA-based encapsulation of different proprioceptive readings (Hip positions, knee force, and current drawn) to determine the appropriate velocity and state of the robot. However, this approach of abstracting results in uncertainty when the extra weight is applied to the robot (\autoref{fig:pca}). Similarly, RL-based methods like~\cite{lee2020learning}, \cite{agarwal2023legged} generalize over different types of legged robots by adapting the various privileged information like simulated terrain profile, friction coefficients, and robot proprioception but they are not well acquainted with the real world terrain representations because of the sim-to-real distinction. 

\subsection{Stability Estimation Across Complex Terrain}\label{stability}
Fundamental physics provides a solid foundation to asses the stability of a physical object. Evaluating physical concepts like the center of gravity, establishing equilibrium, and the moment of force acting on the body determines the state. But with legged robots, on volatile surfaces, it's more complicated. Robots like Boston Dynamic's spot acquire state-of-the-art SDK to stabilize their state. However, we have learning-oriented literature that utilizes various stability modules from the robot's proprioception for supervision to either estimate the terrain safety or navigate the robot more robustly. Works like~\cite{elnoor2024pronav}, provide resistance to the robot's motion using proprioceptive readings as the vibration cost which could determine the stability of the robot. On the other hand, \cite{weerakoon2024vapor} uses direct proprioceptive measures to regulate between a holonomic and non-holonomic action space to reduce the risk of entrapment. 
Some of the quantifications rely on a more detailed analysis~\cite{cai2024evora} of the proprioceptive parameters. On the other hand, simulated observations~\cite{agarwal2023legged}, \cite{lee2020learning} tend to be useful in regulating the robot's locomotion. In contrast, we incorporate the tradeoff between the slips and the velocity acquired by the robot when traversed over different terrains. 
\subsection{Payload Adaptation}
Classical mechanics provide ample evidence to provide payload adaptation to a mobile robot. But most of the parameters are non-differentiable and often rely on the CoM(Center of mass) based stability of a robot. Therefore,~\cite{ding2020locomotion} introduces a CoM estimation for quadruped with varying payloads using Model Predictive Control. Another adaptive PID controller that is robust towards varying payloads~\cite{lee2020adaptive} uses a time delay control model to estimate the gains influenced by varying payloads. 
In contrast, we tend to utilize the proprioception from the movement of the legged robot to significantly adapt different payloads.\vspace{-0.2cm}
\subsection{Pointer Networks for Sequence Learning}\label{pointer}
Pointers networks~\cite{vinyals2015pointer} have been widely used in many combinatorial problems~\cite{ma2019combinatorial}, \cite{velivckovic2020pointer}, \cite{sun2019divgraphpointer}. One major application of utilizing sequential learning is in LLMs for text generation and understanding~\cite{chen2019distilling}, \cite{yan2021unified}. In a conventional text generation using pointer network-based PLM (Pre-trained language models)~\cite{hu2023survey} generate \textit{segments} of the words of interest sequentially. Given the input \textit{x}, the output sequence \textit{y} consists of index numbers corresponding to the positions of words in \textit{x}. The indices in \textit{y} represent class labels which are further processed as vector embeddings and determine the correct sequence of indices in a given sentence. Another example~\cite{vinyals2019grandmaster} incorporates pointer networks in context learning for linear functions (i.e. Text to Prompt). In the context of this work, we adapt sequential learning to establish context between the exteroceptive measures concerning the robot's movement(Section~\ref{Seq}).  

\section{Design}
\label{sec:approach}
This section details our network architecture, learning process, and implementation details. We also discuss the learning mechanism introduced in this work along with the potential benefits of incorporating pointer networks in establishing a low-level representation of terrain estimation and understanding. The overall flowchart of our training pipeline can be seen in \autoref{overview}.


\subsection{Architecture}
\subsubsection{Sequential Learning for Temporal Coherence}\label{Seq} 
Our architectural design is inspired by the pointer networks as mentioned in~\autoref{pointer}. The weight-sharing configuration in a pointer network allows us to learn the temporal relationship between the images being seen and their corresponding proprioception (joint forces and foot movements).
The system collects data from multiple sensor streams (camera, proprioceptive sensors, odometry) and synchronizes them to form multiple sequences. A sequence can be defined as a tuple of synchronized sensor data at a given time \( t \). \textbf{Exteroception}: \( \mathbf{I}_t \in \mathbb{R}^{H \times W \times 3} \) \textbf{Proprioception}: \( \mathbf{P}_t \in \mathbb{R}^{d_P} \) \textbf{Odometry data}: \( v_{\text{applied}, t} \in \mathbb{R} \)

For each time \( t \), a sequence is formed as:
\[
S_t = (\mathbf{I}_t, \mathbf{P}_t, v_{\text{applied}, t})
\]
where \( S_t \) represents the synchronized data for time \( t \).

\begin{figure}
    \centering
    \includegraphics[width=8cm,height=4.4cm]{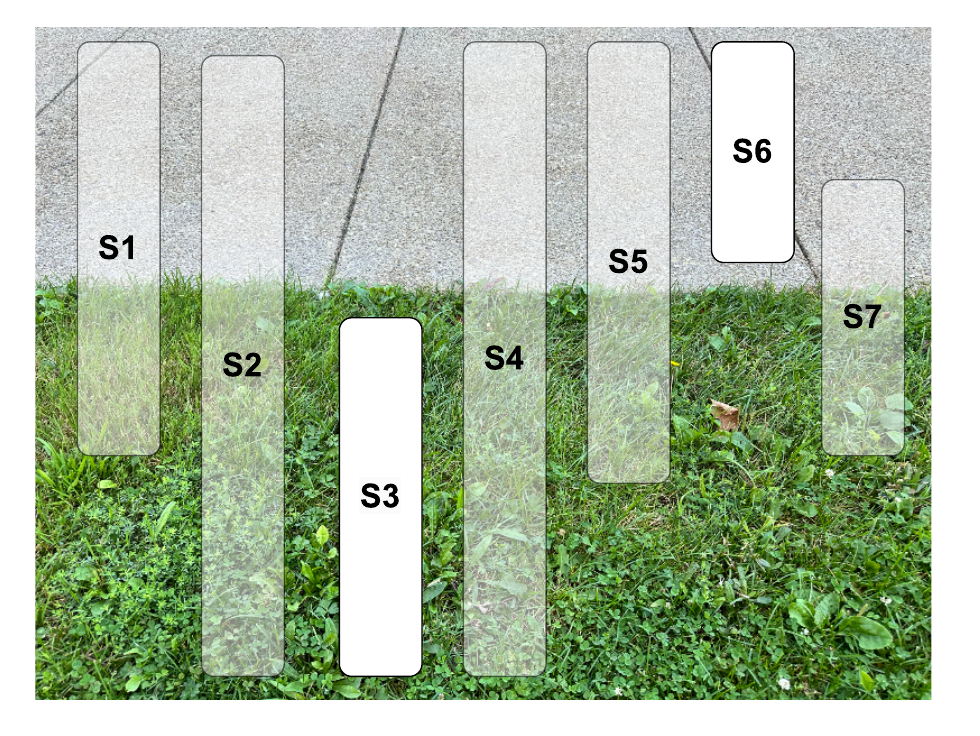}
    \caption{ Sequences S3 and S6 selected from the context vector $\mathbf{c}_t$ based on their high confidence values helps in adapting high-level representation of specific terrains (grass and concrete).} \vspace{-0.5cm}
    \label{fig:transitions}
\end{figure}

The sequences of different sizes are randomly shuffled and divided into mini-batches for learning. If the mini-batch size is \( B \), a mini-batch at from a random iteration \( i \) can be represented as:
\[
B_i = \{ S_{t_1}, S_{t_2}, \dots, S_{t_n} \}
\]
Each batch contains \( n \) sequences, wherein each sequence \( S_t \) consists of:
\( \mathbf{I}_t \) \&
\( \mathbf{P}_t \)
with varying velocities \( v_{\text{applied}}\) at time $t$. These velocities are completely random and considered \textit{weak labels} for a particular sequence during the data collection. These batches with random sequences are used to establish temporal relationships with certain terrain types and their most suited proprioception. Fig.~\ref{fig:transitions} represents the trivial set of selected sequences during the training with a mini-batch.

\begin{figure*}
    \centering
    \includegraphics[width=15cm, height=5.7cm]{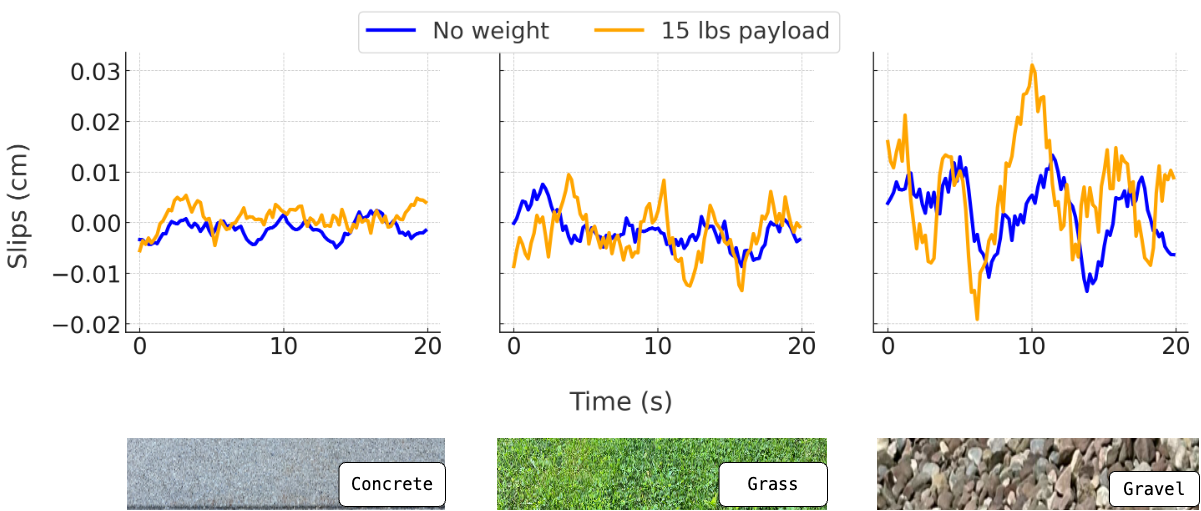}
    \caption{Raw foot slips from the legged robot's proprioception data $\mathbf{P}_t \in \mathbb{R}^{d_P}$ used in PANOS to supervise the training module. We can infer the variability of the robot's foot slips from the body frame (0 as a reference) occurring on different terrains w and w/o payload.}
    \label{fig:raw}
\end{figure*}

\textbf{Raw Proprioception Unfolding: $\mathbb{R}^{d_P}$
} 
\begin{itemize}
    \item \textbf{Joints:} Velocity \& Effort
    \item \textbf{Hips:} Position \& Velocity
    \item \textbf{Feet/Slips:} Position \& Velocity (Fig.~\ref{fig:raw})
\end{itemize}

\subsubsection{Proprioceptive and Exteroceptive Encoding} \label{encodings}
For each \( S_t \), an image \( \mathbf{I}_t \) is passed through the pre-trained vision transformer DINO~\cite{oquab2023dinov2}, which outputs visual tokens:
\[
\mathbf{F}_t^{\text{visual}} = \text{ViT}(\mathbf{I}_t) \in \mathbb{R}^{n_v \times d_v}
\]
Here, \( \mathbf{F}_t^{\text{visual}} \) are the visual tokens for the sequence at time \( t \), with \( n_v \) being the number of residual tokens and \( d_v \) is the token dimension. Similarly, The proprioceptive data \( \mathbf{P}_t \) is passed through a fully connected neural network the \textit{Proprioceptive Encoder} to produce encoded proprioceptive features of the same size as \textbf{P}:
\[
\mathbf{F}_t^{\text{proprio}} = \text{ProprioEncoder}(\mathbf{P}_t) \in \mathbb{R}^{d_P}
\]
Notably, we avoid any level of abstraction in proprioceptive readings to let each feature represent visual tokens of different terrain types and payload adaptation can be made adaptive by using the encoded $\textbf{P}$ transformations across different terrains.

\subsubsection{Intermediate confidence sharing using Pointer Network} \label{confidence}
We have a neural network that takes the encodings and learns to establish a context-aware relationship between the sequences using visual tokens-proprioceptive encodings obtained from \autoref{encodings}.
The context vector is then computed as the weighted sum of the visual tokens and the attention weights:
\[
\mathbf{c}_t = \sum_{i=1}^{n_v} a_t^{(i)} \mathbf{F}_t^{\text{visual}, (i)} \in \mathbb{R}^{d_v}
\]
This context vector encapsulates the visual information most relevant to the current proprioceptive state.

The confidence score for each sequence is computed based on a predefined metric that measures how reliable the current proprioceptive state is for predicting velocity. 
\[
\text{Confidence}_t = 1 - \text{slip}_t
\]
where \( \text{slip}_t \) is the average foot slip, and a lower slip value corresponds to a higher confidence score. From \autoref{fig:raw}, we can notice the impact on slips being highly distant in different terrain types.

\subsubsection{Context-Aware Velocity Regulation}

The context vector $C_t$ obtained represents the weighted confidence representing terrain-proprioception pairs. This vector goes to the MLP module which predicts the velocity component using the learned $\mathbf{c}_t $ selection. The final velocity propagation is:

\[\hat{v}_{t} = \text{MLP} \left( \underset{S_t}{\arg\max} \left( \sum_{i=1}^{n_v} a_{t}^{(i)} F_{\text{visual}, t}^{(i)} \cdot \text{Confidence}_t \right) \right)\]

The velocity $\hat{v}_{t}$ generated from our network serves as the optimal input to the robot in order to maintain good stability while traversing on irregular terrains with heavy payload on the body. This is elaborated further in our evaluation Section~\ref{sec:evaluation}.

\subsection{Losses}
The primary goal is to minimize the slip parameters (Eq.~\ref{eq3}) while achieving the maximum velocity acquired within a sequence during the data collection. The loss minimization $\mathcal{L}$ is over each sequence:
For each mini-batch \( \mathcal{B}_i \), the predicted velocities \( \hat{v}_t \) are compared to the applied velocities (Eq~\ref{eq5}) \( v_{\text{applied}, t} \), and an MSE loss~(Eq.\ref{eq2}) is computed for each sequence $\mathbf{S}_t$. The total loss function is a combination of \textit{Velocity Loss} and \textit{Slip Penalty} with a learnable parameter $\alpha$ as shown in Eq.~\ref{eq4}. In our loss implementation, we restrict the total loss from being negative due to the high velocities applied during the data recording.

\begin{equation}\label{eq2}
\mathcal{L}_{\text{velocity}} = \frac{1}{N} \sum_{i=1}^{N} \left( \hat{v}_i - v_{\text{applied}, i} \right)^2
\end{equation}

\begin{equation}\label{eq3}
\mathcal{L}_{\text{slip}} = \frac{1}{M} \sum_{j=1}^{M} \text{slip}_j
\end{equation}

\begin{equation}\label{eq4}
\mathcal{L}_{\text{total}} = \mathcal{L}_{\text{velocity}} - \alpha \cdot \mathcal{L}_{\text{slip}}
\end{equation}

\begin{equation}\label{eq5}
v_{\text{applied}} = \text{Odometry}(t)
\end{equation}

\section{Evaluation}\label{sec:evaluation}

\begin{figure*}
    \centering
    \includegraphics[width=15cm, height=5.8cm]{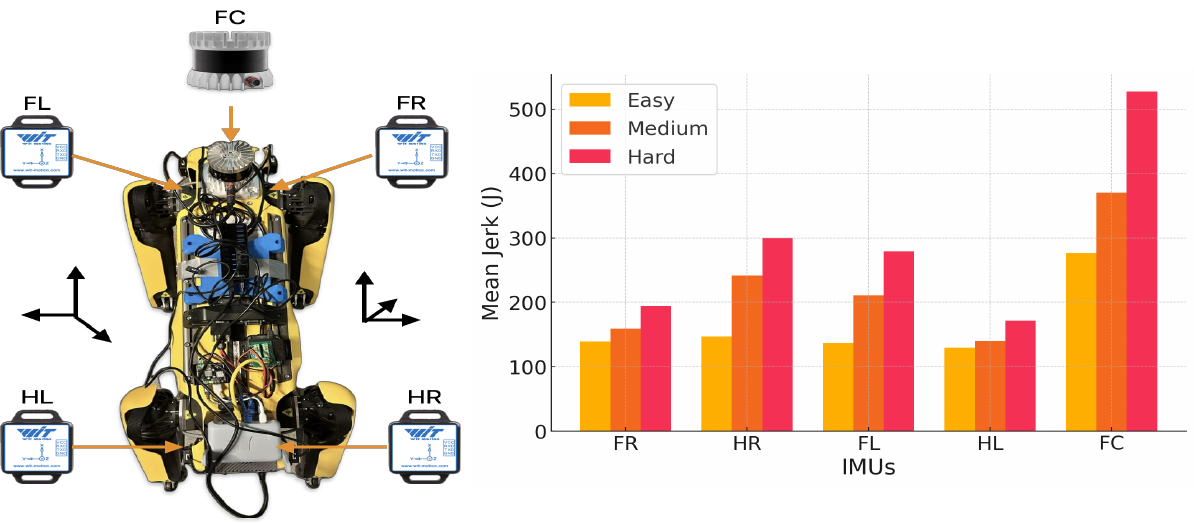}
    \caption{Stability Modeling: Setup of IMU configuration used to measure the stability of different types of terrain. As a modeling parameter, we define the reliability of the setup by measuring the mean jerk $\bar{J}$ on three different terrains East(concrete), Medium(grass), and Hard (Gravel). The graph shown above shows the $\bar{J}$ acting on the five IMUs used in the setup while driving on three different terrains with three distinct properties.}
    \label{fig:imus}
\end{figure*}

\textbf{Data Collection and Training:} Data for training was collected across different terrains like grass, concrete, gravel, pebble sidewalks, etc. During the data collection, we recorded 4392 images (1920 x 1080), ~14k ROS messages from the spot's joints (position and effort), hips (position and velocity), feet (slips), and the robot's velocity. During the data collection, we leverage the adaptability of our approach to train the algorithm with incomplete and ambiguous labels (i.e. desired velocities). The velocity labels within each sequence $S_t$ are considered  \textit{weak} and are only considered for training if they establish a reasonable tradeoff in increasing confidence (~\autoref{confidence}). These \textit{weak} labels are recorded when the robot is manually controlled with varying velocities over multiple terrains. 
Figure~\ref{fig:imus} shows the complete setup used to collect data for training and testing. We use Boston Dynamics Spot with a RealSense D435 camera mounted at a downward angle at the bottom of the robot's face, 4 WitMotion IMUs placed at each corner of the robot's body, and an Ouster OS-1 Lidar with IMU at the front center of the robot. The five IMUs on the body are used to quantify the stability (\autoref{sec:4d}). The network is trained on a Desktop with an NVIDIA RTX A6000 GPU and inference is run online on an NVIDIA Jetson AGX Orin mounted on the robot and controls it.

\textbf{Testing:} 
To test PANOS, we do two runs of each (with and without payload) using our trained model on an unseen environment with multiple terrain variations as seen in \autoref{fig:firstpage_intro}. During the test, we used multiple IMUs (\autoref{fig:imus}) to determine the stability of our robot. We do a live test by commanding the robot from a dedicated source to the destination using the velocities from PANOS and various baselines. We compare various baselines that estimate the desired velocities using exteroception~\cite{sathyamoorthy2023vern} or a combination of exteroception and proprioception~\cite{weerakoon2024vapor}. We also test the spot's inbuilt mechanism~\cite{bostondynamics2023} that helps it adapt the gait/velocity to maintain stability. We keep the payload estimation, obstacle avoidance, test terrain, source and destination waypoints, and other stability modules from the spot's SDK intact and consistent to perform a fair comparison. The initial velocity (2m/s) used for testing remains the same for all the baselines and PANOS. The description of the baselines used in this evaluation is as follows:

\begin{itemize}
    \item \textbf{Spot's Inbuilt}~\cite{bostondynamics2023}: The internal feedback system of Spot is equipped with various IMUs to monitor and regulate the acceleration and stability of the robot. The real-time feedback loop from the robot's internal software helps it to adjust the 3d acceleration even during a manual input of velocities. 
    
    \item \textbf{VERN}~\cite{sathyamoorthy2023vern}: Vegetation-aware navigation method that computes the robot's velocity using exeroception (images) to safely traverse in real-world unstructured environments. VERN provides velocity estimation using a local cost map generated by the system to determine the pliability of a terrain.

    \item \textbf{VAPOR}~\cite{weerakoon2024vapor}: Reinforcement Learning (RL) method to provide context-aware feedback to regulate velocities and switch the robot's state motion between holonomic and non-holonomic. VAPOR trains a novel RL policy perceiving exteroception (3D LiDAR) and proprioception (robot's joints) to regulate the velocity while traversing complex terrains. 
\end{itemize}

\begin{table*}
\centering
\resizebox{\textwidth}{!}{%
\begin{tabular}{@{}l|llllll|llllll@{}}
\toprule
\multicolumn{1}{c|}{\multirow{2}{*}{Method}} & \multicolumn{6}{c|}{W/O Payload}                          & \multicolumn{6}{c}{W Payload}         \\ \cmidrule(l){2-13}
\multicolumn{1}{c|}{}                         & $\bar{J}_{FR}$ (\( \text{m/s}^3 \)) & $\bar{J}_{FL}$ (\( \text{m/s}^3 \))& $\bar{J}_{HR}$ (\( \text{m/s}^3 \)) & $\bar{J}_{HL}$ (\( \text{m/s}^3 \)) & $\bar{J}_{HC}$ (\( \text{m/s}^3 \)) & Cost (cm)& $\bar{J}_{FR}$ (\( \text{m/s}^3 \))& $\bar{J}_{FL}$ (\( \text{m/s}^3 \))& $\bar{J}_{HR}$ (\( \text{m/s}^3 \))& $\bar{J}_{HL}$ (\( \text{m/s}^3 \))& $\bar{J}_{HC}$ (\( \text{m/s}^3 \))& Cost (cm)\\ \midrule
Spot's Inbuilt~\cite{bostondynamics2023}   & 240.01  & 792.87 & 354.63 & 739.26  & 546.95 & 0.93  & 341.54 & 779.88 &925.05 &599.65&836.44 & 3.49        \\
 \midrule
VAPOR~\cite{weerakoon2024vapor}   & 255.22   & 667.44 & 333.5 & 653.49  & 449.94 & 0.95 & 283.94 & 648.49 & 773.94 & 449.55 & 749.40 & 4.04        \\
 \midrule
VERN!\cite{sathyamoorthy2023vern}   & 262.34   & 645.66 & 304.93 & 584.39  & 394.59 & 1.1 & 302.39 & 449.10 & 610.45 & 503.30 & 798.39 & 3.77        \\
 \midrule
PANOS & \textbf{221.17}   & \textbf{227.32} & \textbf{208.93} & \textbf{440.45}  & \textbf{386.44} &0.93  & \textbf{195.20} & \textbf{249.67} &\textbf{315.80} & \textbf{207.01} & \textbf{365.17} & \textbf{3.26}                     
      \\ \bottomrule
\end{tabular}%
}
\caption{Evaluation of different methods using mean jerk ($\bar{J}$) on each imu (FR, FL, HR, HC, and FC) mounted on the robot. We also evaluate our approach with a pre-defined vibration cost which signifies the offset of Hip movements from the body frame.}
\label{tab:vo-eval}
\end{table*}

\subsection{Evaluation Metrics}
\subsubsection{Vibration Cost}\label{Vibration Cost}
As used in~\cite{weerakoon2024vapor, elnoor2024pronav}, 
 we also adapt the vibration cost that determines the overall offset of the hip joints from the body as a metric. Vibration costs determine the offset of the robot's joints from the body frame. While testing our approach on varying terrains, we extract the joint offsets (positions) from \( \mathbf{P}_t \in \mathbb{R}^{d_P} \). The obtained cost acts as a measure to determine the robot's stability.
 

\label{sec:result}
\subsubsection{Stability Modeling}\label{sec:4d}
To more precisely measure the stability of our robot, we use an IMU-based setup in our experiments. 
We instrument five different IMUs (see \autoref{fig:imus}) mounted on each corner (Front Right, Front Left, Hind Right, Hind Left) and center (Front Center) of the robot's body. 
To estimate the stability, we define the mean magnitude of jerks acting across all five IMUs:
\[
J_i(t) = \sqrt{\left(\frac{da_{x,i}(t)}{dt}\right)^2 + \left(\frac{da_{y,i}(t)}{dt}\right)^2 + \left(\frac{da_{z,i}(t)}{dt}\right)^2}
\]

where \(a_{x,i}(t)\), \(a_{y,i}(t)\), and \(a_{z,i}(t)\) are the acceleration components from each IMU \(i\) at time \(t\). To normalize the axis we take the magnitude of all derivatives \(\frac{da}{dt}\).

Finally, the mean magnitude of jerks across all IMUs is: \vspace{-0.3cm}

\[
\text{Mean Jerk} (\bar{J}) = \frac{1}{5} \sum_{i=1}^{5} J_i(t)
\]

To estimate the percentage improvement in jerk is calculated as: $\frac{\bar{J}(\text{baseline}) - \bar{J}(\text{PANOS})}{\bar{J}(\text{baseline})}$.

\subsection{Results and Analysis}
We implement PANOS and other baselines on an offroad scenario with different variations in terrains. Figure.~\ref{fig:firstpage_intro} shows the test scenario we used to perform our experiments. We did two runs per evaluation for both with payload (15 lbs) and without payload (except the sensor weights of around 1kg). 

\textbf{Stability:} From~\autoref{tab:vo-eval}, we can observe the performance of PANOS in reducing the mean jerk $\bar{J}$ for both with and without payload runs. The overall reduction of instability obtained from PANOS is up to \textbf{53}\% w payload and \textbf{44}\% w/o payload. By learning from both exteroception as well as proprioception, PANOS is able to generate velocities that reduce instability and allow for stable locomotion. Note that changing payloads does not need any retraining. 

\textbf{Vibration Cost:} Vibration is the deviation of leg movement from the body frame (\autoref{Vibration Cost}). From \autoref{tab:vo-eval}, PANOS reduces vibration cost by up to \textbf{20\%} compared to the baselines. 

\section{Conclusion and Future Work}\label{sec:conclusion}
We present PANOS, a weakly supervised approach for payload-aware navigation for quadruped robots in irregular and complex terrains. Our approach incorporates the exteroception and proprioception from the robot to estimate the desired velocity.
Our approach not only accounts for the stability of the robot while navigating on irregular terrains but also self-tunes itself when the robot is under heavy payload. Our learning framework without any desired label incorporates the visual representations of specific terrain and creates a correspondence with the appropriate velocity required for specific terrain. We evaluate our method in terms of stability and leg-joint vibrations impacting the robot's body. We observe our method to acquire more stability up to \textbf{53\%} when tested on an irregular terrain with varying payloads. Our method also reduces the vibrations from the leg of a robot up to \textbf{20\%}. In future work, we would aim to integrate other sensing modules like gait adaptation for a robot to change its state according to the terrain type. 
\newpage
\bibliographystyle{IEEEtran}
\bibliography{references}
\end{document}